\documentclass{article}
\usepackage{spconf,amsmath,graphicx}

\usepackage{algorithm}
\usepackage{algorithmic}

\usepackage{booktabs}  

\usepackage{refcount}
\usepackage{bbm}
\usepackage{makecell}
\usepackage{amssymb}
\usepackage{amsthm}
\usepackage{multirow}
\usepackage{xspace}
\usepackage{subcaption}

\usepackage{newfloat}
\usepackage{listings}
\floatstyle{ruled}
\newfloat{listing}{tb}{lst}{}
\floatname{listing}{Listing}

\usepackage{soul}
\usepackage{tikz}
\usetikzlibrary{calc}
\usetikzlibrary{decorations.pathmorphing}

\definecolor{citecolor}{RGB}{34,139,34}
\usepackage[pagebackref=true,breaklinks=true,letterpaper=true,colorlinks,citecolor=citecolor,bookmarks=false]{hyperref}

\usepackage{xspace}

\def\@onedot{\ifx\@let@token.\else.\null\fi\xspace}
\DeclareRobustCommand\onedot{\futurelet\@let@token\@onedot}

\newcommand{\X}{\cal X}
\newcommand{\Y}{\cal Y}
\newcommand{\ger}{g_{w_{\text{er}}}}
\newcommand{\gid}{g_{w_{\text{id}}}}

\renewcommand{\[}{\begin{eqnarray}}
\renewcommand{\]}{\end{eqnarray}}
\newcommand{\R}{\mathbb{R}}

\title{Speaker normalization for self-supervised \\ speech emotion recognition}
\name{Itai Gat, Hagai Aronowitz, Weizhong Zhu, Edmilson Morais, Ron Hoory}
\address{IBM Research AI}
\begin{document}
%
\maketitle
\begin{abstract}
     Large speech emotion recognition datasets are hard to obtain, and small datasets may contain biases. Deep-net-based classifiers, in turn, are prone to exploit those biases and find shortcuts such as speaker characteristics. These shortcuts usually harm a model's ability to generalize. To address this challenge, we propose a gradient-based adversary learning framework that learns a speech emotion recognition task while normalizing speaker characteristics from the feature representation. We demonstrate the efficacy of our method on both speaker-independent and speaker-dependent settings and obtain new state-of-the-art results on the challenging IEMOCAP dataset.
\end{abstract}
\section{Introduction}

Over the last decade, advances in speech processing have been extremely impressive. Challenges like personal voice assistants that seemed daunting merely ten years ago are now part of our day-to-day routine. A crucial part of those systems is to interact with the user and  understand his intents and emotions. Understanding the user's emotion can build trust between the user and the system and improve the user experience. Improving the accuracy of speech emotion recognition has, therefore, been a major research focus in recent years.

Reported improvements are, to a large extent, due to the availability of large general purpose and task-specific datasets~\cite{voxceleb, librispeech}, computational performance advances (e.g., for GPUs), and a better understanding of how to encode inductive biases into deep neural nets. The use of self-supervised learning methods has played a crucial role in the advancements of textual and visual modalities due to their ability to learn strong feature representations from unlabeled data. Recently, these techniques have shown success in the speech processing community as well.

The annotation of speech emotion recognition is challenging, because it is affected by the annotator's bias towards linguistic, cultural, and social constraints. Speech emotion recognition, in particular, yields biased spurious correlations between speaker characteristics and the emotional class of the recording. Several datasets have been created over the years for the training and evaluation of machine learning methods for speech emotion recognition~\cite{iemocap, RAVDESS, savee}. Most of them are relatively small. The difficulty in learning from biased and small datasets are the main challenges facing the deep learning research community.

\begin{figure*}[t!]
\centering
\includegraphics[width=\textwidth]{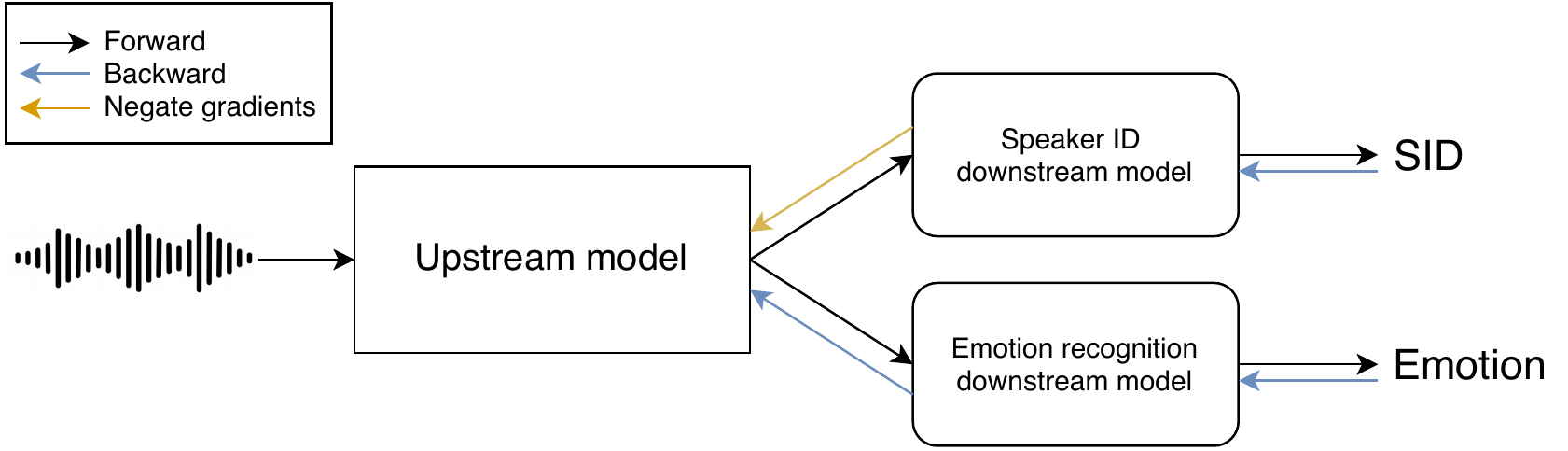}
\caption{The architecture of our method: In the forward pass, we learn both speaker identification (SID) and emotion tasks. Speaker features enable a good fit to the training dataset. However, they hinder generalization for unknown speakers. To improve generalization for an unknown speaker, we propose disentangling speaker features from the upstream representation by negating the gradients from the downstream model back to the input in the backward pass.}
\label{fig:method}
\end{figure*}

The classic self-supervised learning process relies on a representation trained on a large unlabeled dataset, and a downstream task trained on a relatively small labeled dataset. Generally, our method enhances a downstream task performance by using a third dataset with labels different from the downstream task labels. For example, in this work, for speaker emotion recognition, our method normalizes undesired characteristics from the self-supervised representation to improve performance on the speech emotion recognition task. We carry this out by learning a feature representation that excels at speech emotion recognition while being robust enough for speaker characteristics (see Fig.~\ref{fig:method}). Our proposed method outperforms the current state-of-the-art results for both speaker-dependent and speaker-independent settings.

In summary, we propose a general framework for speaker characteristics normalization from a self-supervised representation. We address the small dataset settings issue and propose a framework for it on the IEMOCAP benchmark. Through extensive experiments, we show that our method outperforms the current speech emotion recognition state-of-the-art results on several setups.

The remainder of the work is organized as follows: Sec.~\ref{sec:related_work} provides an overview of related work. Sec.~\ref{sec:method} describes our proposed method. Sec.~\ref{sec:experiments} reports the experiments and results of our method. Finally, Sec.~\ref{sec:conclusions} concludes the paper.

\section{Related work}\label{sec:related_work}

\subsection{Self-supervised trained models}

The field of deep learning research has significantly benefited from self-supervised learning. In this paradigm, a task-independent model is pre-trained using large volumes of unlabeled data, and a task-specific model is  trained over the self-supervised model. The majority of self-supervision techniques in speech processing fall into three categories. In the first category, different architectures combined with a constructive InfoNCE loss are used~\cite{oord2018representation, cpc, modified-cpc, wav2vec, wav2vec2}. The second category is based on masked token classification~\cite{hubert}. The third category employs reconstruction loss using various techniques such as generation of future frames~\cite{apc, vq-apc} and the encoder-decoder approach to reconstruct masked parts of the input~\cite{mockingjay, tera, npc}.

\subsection{Feature normalization}

Deep neural nets excel at fitting with the training set. As a consequence, they tend to heavily leverage superficial correlations between the data features and the labels. While this enhances performance on the training set, it may harm the generalization capabilities of the model. To overcome this, various methods propose to eliminate the model's ability to learn undesirable cues. Nagrani et al.~\cite{nagrani2020disentangled} suggest using a "confusion loss," which is a cross-entropy loss computed by comparing the prediction to a uniform distribution. Ganin et al.~\cite{ganin2015unsupervised} use extra knowledge regarding the data-domain to tackle a domain adaptation problem. They propose to normalize domain features by negating gradients of a loss that predict the domain label. In contrast to those methods, we normalize cues based on a task rather than a domain. Additionally, our method focuses on the self-supervised representation framework.

\subsection{Emotion recognition}

Speech emotion recognition predicts an emotion based on an utterance. The most widely used benchmark for speech emotion recognition is the interactive emotional dyadic motion capture database (IEMOCAP)~\cite{iemocap}. For the purpose of speech emotion recognition, early end-to-end methods combine a convolutional neural network (CNN) and a long short-term memory (LSTM)~\cite{Keren2016ConvolutionalRA, Mirsamadi2017AutomaticSE, Trigeorgis2016AdieuFE}. Later, attention-based models outperformed the CNN and LSTM combination due to their ability to focus on specific parts of the input~\cite{moine2021speaker, Chen20183DCR, Li2019ImprovedES}. In recent years, self-supervised learning models have generated significant interest in speech processing research due to their ability to learn high-quality representations from unlabeled data. Reflecting this, Yang et al.~\cite{s3prl} demonstrate in their benchmark that self-supervised models produce state-of-the-art results in emotion recognition. We present a method for enhancing speech emotion recognition by combining self-supervised models with the normalization of speaker characteristics.


\section{Method}\label{sec:method}
Learning a discriminative model amounts to fit function $f_w: \X \rightarrow \Y$ from the space of data $x \in \X$ to the space of labels $y \in \Y$ using parameters $w$. The parameters of the function $f_w(x)$ are learned by fitting to a training data $\{(x_1,y_1),..., (x_m,y_m)\}$ using a loss function $\ell(f_w(x_i),y_i)$. Recent works suggest an upstream-downstream architecture where a discriminative learner $f_w$  is composed of a pre-trained upstream learner $h_{w_1}:\mathcal{X} \rightarrow \R^k$ that maps a data point to an embedding representation in $\R^{k}$. Then, a downstream learner $g_{w_2}: \R^{k}\rightarrow\Y$ maps the output of $h_{w_1}$ to the label space $\Y$. This architecture can also be viewed as a composition of $h_{w_1}$ and $g_{w_2}$, i.e., $f_w(x) = (g_{w_2} \circ h_{w_1})(x)$.

A discriminative learning algorithm searches for parameters $w$ to best fit the relation of $(x_i,y_i)$ in the training data. However, it can sometimes exploit undesired cues. For example, in speech emotion recognition, we seek to encourage the learner to ignore speaker characteristics such as gender or any other speaker-specific features. In the following, we propose an approach for learning a task while normalizing cues from a different task (possibly from another dataset) in an upstream-downstream architecture.

\subsection{Speaker normalization}

The upstream-downstream approach allows us to learn more than one task above a single upstream model. An example of this is solving both speaker identification and emotion recognition tasks. In our method, we consider three discriminative learners. The first is an upstream model $h_w$, the second is an emotion recognition learner $\ger$, and the third is a speaker identification classifier $\gid$. To prevent the emotion downstream task from leveraging undesired speaker characteristics, we suggest to normalize them from the upstream representation so that the emotion classifier is unable to leverage those cues. We propose doing this by negating the gradients of the upstream model with respect to the speaker identification task. For the sake of simplicity, we describe our method using stochastic gradient descent (SGD), but it can easily be applied to any optimizer.

Our approach consists of two steps. The first step is a standard gradient-based optimization. For example, in standard gradient-based learning using the SGD algorithm, the upstream and emotion downstream weights update step is 
\[
    w \leftarrow w - \eta\frac{\partial \ell_{\text{er}}\big((\ger \circ h_{w})(x_i),y_i \big)}{\partial w} \\
    w_{\text{er}} \leftarrow w_{\text{er}} - \eta\frac{\partial \ell_{\text{er}}\big((\ger \circ h_{w})(x_i),y_i\big)}{\partial w_{\text{er}}},
\]
where $\eta$ is a learning rate and $\ell_{\text{er}}$ is an emotion recognition loss, e.g., cross-entropy loss. In the second step, we normalize speaker id features from the upstream model by
\[
    w \leftarrow w + \lambda\frac{\partial \ell_{\text{id}}\big((\gid \circ h_{w})(x_j),y_{j}\big)}{\partial w}\label{eq:id_dis} \\
    w_{\text{id}} \leftarrow w_{\text{id}} - \eta\frac{\partial \ell_{\text{id}}\big((\gid \circ h_{w})(x_j),y_{j}\big)}{\partial w_{\text{id}}},
\]
where $\lambda$ is a hyperparameter that sets to which extent we want to normalize speaker features from the upstream model by performing a gradient ascent step on the upstream model, with respect to the speaker id loss $\ell_{\text{id}}$. We illustrate our method in Fig.~\ref{fig:method}. It is important to note that the steps are independent, so the data used for emotion recognition does not need to be labeled with speaker identification, and vice versa. Next, we present two approaches for training our method.

\subsection{Training strategies}\label{sec:strategies}

Self-supervised upstream models often have many parameters. For example, in the HuBERT Large model~\cite{hubert}, there are 317 million parameters, and in HuBERT X-Large~\cite{hubert}, there are nearly a billion parameters. Thus, fine-tuning such networks can be computationally hard. We propose two training procedures:
\begin{enumerate}
    \item Speaker normalization projector: We introduce a new non-linear layer with parameters $\widehat{w}$, which we use as a gate between the upstream and the downstream model. In the emotion recognition step, we add $\widehat{w}$ to the optimization of the upstream model. However, in the speaker ID task, we modify Eq.~\ref{eq:id_dis} to optimize solely $\widehat{w}$. In both tasks, we do not change the optimization of the downstream models. This allows us to skip the upstream's optimization step, which spares the gradients computation overhead in the speaker ID step.
    \item Train all parameters: In this approach, we train the parameters of both upstream and downstream models according to what we describe above. 
\end{enumerate}

In the following section, we discuss both training strategies on speaker-independent and speaker-dependent settings using various training set sizes. We present improvements from state-of-the-art using both training strategies.

\begin{table}[t!]
    \centering
    \begin{tabular}{lccc}
        \toprule
        Method  & 5-fold & SD & AUC$^{*}$\\
        \hline
        MDRE~\cite{yoon2018multimodal} & - & 71.8 & - \\
        DS-LSTM~\cite{wang2020speech} & - & 72.7 & - \\
        MOMA~\cite{moma} & - & 74.8 & - \\
        WISE~\cite{wise} & 66.5 & - & - \\
        wav2vec2-PT~\cite{pepino21_interspeech} & 67.2 & - & - \\
        ACTC~\cite{actc} & 69.0 & - & - \\
        AttPool~\cite{li18c_interspeech} & 71.7 & - & - \\
        \hline
        HuBERT Base~\cite{hubert, s3prl} & 68.9 & 71.2 & 61.0 \\
        HuBERT Base + Ours (SNP) & 69.0 & 72.8 & 61.6\\
        HuBERT Base + Ours (TAP)  & 69.9 & 72.4 & 62.7\\
        \hline
        HuBERT Large~\cite{hubert, s3prl} & 71.9 & 78.3 & 63.9 \\
        HuBERT Large + Ours (SNP)  & 73.9 & 79.9 & 64.2 \\
        HuBERT Large + Ours (TAP)  & \textbf{74.2} & \textbf{81.0} & \textbf{64.9}\\
        \bottomrule
    \end{tabular}
    \caption{State-of-the-art results on IEMOCAP for speaker-independent settings using only audio features for five-fold cross-validation and speaker-dependent settings with a random train-test split (SD). For 5-fold and SD settings, we report the weighted accuracy (WA) metric. Speaker normalization projector (SNP) and train all parameters (TAP) refers to the training strategies described in Sec.~\ref{sec:strategies}. Using our method, we are able to outperform the current speaker-independent state-of-the-art by 2.3\% and speaker-dependent state-of-the-art by 0.5\%. (*) Additionally, we present the AUC for low-resource settings (see Sec.~\ref{sec:low_resource_experiment}), where our approach significantly enhances both the HuBERT Base and Large models.}
    \label{tab:results}
    \vspace{-0.4cm}
\end{table}

\section{Experiments}\label{sec:experiments}

This section presents our approach and compares it to previous work. First, we describe our experimental setup (see Sec.~\ref{sec:experimental}). We then show that our approach outperformed the current state-of-the-art results on the IEMOCAP benchmark (see Sec.~\ref{sec:emotion_results}). We also present a low-resource evaluation setup and show the efficacy of our method on it (see Sec.~\ref{sec:low_resource_experiment}).

\subsection{Experimental setup}\label{sec:experimental}

We performed our experiments on the IEMOCAP dataset~\cite{iemocap}. According to the traditional evaluation process, we used balanced classes: neutral, happy, sad, and angry. There are ten different speakers in the dataset (five female and five male). For the speaker identification task, we used the VoxCeleb dataset~\cite{voxceleb}, which is annotated with $1,251$ speakers and more than $100,000$ utterances. For the upstream model, we used both the HuBERT base and large models~\cite{hubert}. For the downstream model, we used a non-linear projection from the temporal dimension of HuBERT. The range of hyperparameters we considered for $\lambda$ is $[0.01, 0.0001]$, although in the end we obtained the best results using $\lambda=0.001$. Accordingly, we used this value of $\lambda$ in all the experiments. For all experiments, we use a single NVIDIA Tesla V100 GPU.

\subsection{Emotion recognition}\label{sec:emotion_results}

Recent studies suggest various evaluation setups for emotion recognition. Those setups can be divided into two categories: speaker-independent and speaker-dependent. For speaker-independent, we performed five-fold cross-validation, where utterances from two speakers are used for testing and utterances from eight other speakers are used for training and validation in each fold. The stopping criteria play a crucial role in speaker out-of-distribution evaluation as well. We report the accuracy of the test set based on the best epoch on the validation set.

In this work, we investigate generalization capabilities to an unknown speaker. Therefore, we focus on the speaker-independent setup. Nevertheless, we show that our method also improves the speaker-dependent setup where the train-test split is random and includes all speakers in both the train and test sets.

In Table~\ref{tab:results}, we present results of our method on both the speaker-independent (five-fold) and speaker-dependent (SD) setups. For speaker-independent, our second training procedure produced a state-of-the-art (SOTA) result, outperforming the current SOTA by 2.3\% absolute. We show that the improvement is consistent with both HuBERT Large and Base. Moreover, our method improves the SOTA results for speaker-dependent settings by 0.5\% absolute.

We further evaluated our method's ability to normalize speaker information from an upstream model. We trained a classifier twice on a fixed upstream (i.e., without fine-tuning the upstream model) HuBERT Large for speaker identification. First, we trained a downstream model on HuBERT before our speaker normalization method. We obtained an accuracy of 60.7\%. We then trained an additional speaker ID downstream model on a fixed HuBERT that was trained with our proposed method. This resulted in 45.9\% accuracy. Thus, as desired, our method harmed the speaker ID features of the upstream model.

\begin{figure}[t!]
    \centering
    \includegraphics[width=.48\textwidth]{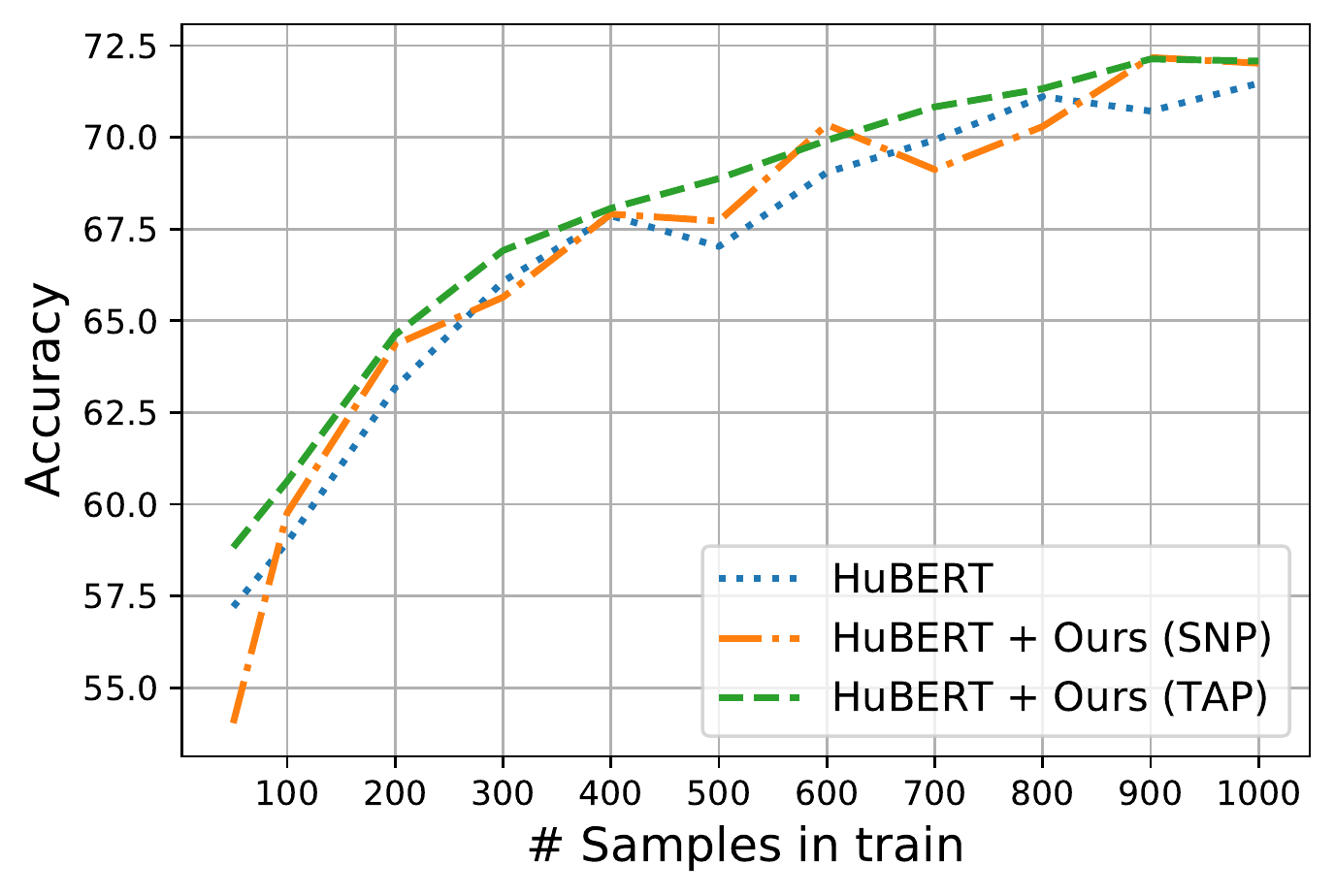}
    \caption{A comparison of our proposed methods with the HuBERT baseline for small data sets: We train classifiers on different amounts of training data per label (x-axis). We observe that the normalization of speaker features from the upstream representation using our method enhances the generalization capabilities of HuBERT.}
    \label{fig:results}
\end{figure}

\subsection{Small data settings}\label{sec:low_resource_experiment}

A high-standard data collection and annotation is often an expensive process. In the following, we suggest quantitative evaluation settings for small data setup.

To test speech emotion recognition with few resources, we propose increasing the number of samples per class in the training set. Then, to stabilize the results, we run each step five times with different random splits and compute each step's mean. Finally, to quantify the overall performance of a given method, we compute the area under the curve (AUC) of Fig.~\ref{fig:results}. Intuitively, the AUC score reflects an average of the scores for each setup we evaluate.

Fig.~\ref{fig:results} presents results for our proposed low-resource setup. In each step, we trained a HuBERT Large model with and without our methods.  We report the AUC of each method in Table~\ref{tab:results}. By using our method, we were able to improve both the Base and Large HuBERT models. In Fig.~\ref{fig:results} we note that our method improves HuBERT accuracy in all settings. This can also be observed in the AUC improvement in Table~\ref{tab:results}.

\vspace{-0.15cm}
\section{Conclusion}\label{sec:conclusions}
\vspace{-0.15cm}

In this paper, we presented a framework for speaker characteristics normalization from a self-supervised feature representation. Our approach combines discriminative learning of a task with adversary learning of another task. Furthermore, our method enables to use of different datasets for each task. We tested it on top of various models and achieved strong state-of-the-art results in speech emotion recognition. We also proposed studying low-resource settings using a modified version of IEMOCAP and showed the success of our method on it.

\clearpage

{\small
\bibliographystyle{IEEEbib}
\bibliography{refs}

\begin{thebibliography}{10}

\bibitem{voxceleb}
Arsha Nagrani, Joon~Son Chung, and Andrew Zisserman,
\newblock ``Voxceleb: a large-scale speaker identification dataset,''
\newblock in {\em Interspeech}, 2017.

\bibitem{librispeech}
Vassil Panayotov, Guoguo Chen, Daniel Povey, and Sanjeev Khudanpur,
\newblock ``Librispeech: An asr corpus based on public domain audio books,''
\newblock in {\em ICASSP}, 2015.

\bibitem{iemocap}
Carlos Busso, Murtaza Bulut, Chi-Chun Lee, Abe Kazemzadeh, Emily Mower, Samuel
  Kim, Jeannette~N Chang, Sungbok Lee, and Shrikanth~S Narayanan,
\newblock ``Iemocap: Interactive emotional dyadic motion capture database,''
\newblock {\em Language resources and evaluation}, 2008.

\bibitem{RAVDESS}
Steven~R. Livingstone and Frank~A. Russo,
\newblock ``The ryerson audio-visual database of emotional speech and song
  (ravdess): A dynamic, multimodal set of facial and vocal expressions in north
  american english,''
\newblock {\em PLOS ONE}, 2018.

\bibitem{savee}
Bogdan Vlasenko, Bjorn Schuller, Andreas Wendemuth, and Gerhard Rigoll,
\newblock ``Combining frame and turn-level information for robust recognition
  of emotions within speech,''
\newblock 2007.

\bibitem{oord2018representation}
Aaron van~den Oord, Yazhe Li, and Oriol Vinyals,
\newblock ``Representation learning with contrastive predictive coding,''
\newblock {\em arXiv:1807.03748}, 2018.

\bibitem{cpc}
Aaron van~den Oord, Yazhe Li, and Oriol Vinyals,
\newblock ``Representation learning with contrastive predictive coding,''
\newblock {\em arXiv:1807.03748}, 2018.

\bibitem{modified-cpc}
Morgane Riviere, Armand Joulin, Pierre-Emmanuel Mazar{\'e}, and Emmanuel
  Dupoux,
\newblock ``Unsupervised pretraining transfers well across languages,''
\newblock in {\em ICASSP}, 2020.

\bibitem{wav2vec}
Steffen Schneider, Alexei Baevski, Ronan Collobert, and Michael Auli,
\newblock ``wav2vec: Unsupervised pre-training for speech recognition,''
\newblock in {\em Interspeech}, 2019.

\bibitem{wav2vec2}
Alexei Baevski, Henry Zhou, Abdelrahman Mohamed, and Michael Auli,
\newblock ``wav2vec 2.0: A framework for self-supervised learning of speech
  representations,''
\newblock in {\em NeurIPS}, 2020.

\bibitem{hubert}
Wei-Ning Hsu, Benjamin Bolte, Yao-Hung~Hubert Tsai, Kushal Lakhotia, Ruslan
  Salakhutdinov, and Abdelrahman Mohamed,
\newblock ``Hubert: Self-supervised speech representation learning by masked
  prediction of hidden units,''
\newblock {\em arXiv:2106.07447}, 2021.

\bibitem{apc}
Yu-An Chung, Wei-Ning Hsu, Hao Tang, and James Glass,
\newblock ``An unsupervised autoregressive model for speech representation
  learning,''
\newblock {\em arXiv:1904.03240}, 2019.

\bibitem{vq-apc}
Yu-An Chung, Hao Tang, and James Glass,
\newblock ``Vector-quantized autoregressive predictive coding,''
\newblock in {\em Interspeech}, 2020.

\bibitem{mockingjay}
Andy~T Liu, Shu-wen Yang, Po-Han Chi, Po-chun Hsu, and Hung-yi Lee,
\newblock ``Mockingjay: Unsupervised speech representation learning with deep
  bidirectional transformer encoders,''
\newblock in {\em ICASSP}, 2020.

\bibitem{tera}
Andy~T Liu, Shang-Wen Li, and Hung-yi Lee,
\newblock ``Tera: Self-supervised learning of transformer encoder
  representation for speech,''
\newblock {\em TASLP}, 2021.

\bibitem{npc}
Alexander~H Liu, Yu-An Chung, and James Glass,
\newblock ``Non-autoregressive predictive coding for learning speech
  representations from local dependencies,''
\newblock {\em arXiv:2011.00406}, 2020.

\bibitem{nagrani2020disentangled}
Arsha Nagrani, Joon~Son Chung, Samuel Albanie, and Andrew Zisserman,
\newblock ``Disentangled speech embeddings using cross-modal
  self-supervision,''
\newblock in {\em ICASSP}, 2020.

\bibitem{ganin2015unsupervised}
Yaroslav Ganin and Victor Lempitsky,
\newblock ``Unsupervised domain adaptation by backpropagation,''
\newblock in {\em ICML}, 2015.

\bibitem{Keren2016ConvolutionalRA}
Gil Keren and Bj{\"o}rn Schuller,
\newblock ``Convolutional rnn: An enhanced model for extracting features from
  sequential data,''
\newblock {\em IJCNN}, 2016.

\bibitem{Mirsamadi2017AutomaticSE}
Seyedmahdad Mirsamadi, Emad Barsoum, and Cha Zhang,
\newblock ``Automatic speech emotion recognition using recurrent neural
  networks with local attention,''
\newblock {\em ICASSP}, 2017.

\bibitem{Trigeorgis2016AdieuFE}
George Trigeorgis, F.~Ringeval, R.~Brueckner, E.~Marchi, Mihalis~A. Nicolaou,
  Bj{\"o}rn Schuller, and S.~Zafeiriou,
\newblock ``Adieu features? end-to-end speech emotion recognition using a deep
  convolutional recurrent network,''
\newblock {\em ICASSP}, 2016.

\bibitem{moine2021speaker}
Cl{\'e}ment~Le Moine, Nicolas Obin, and Axel Roebel,
\newblock ``Speaker attentive speech emotion recognition,''
\newblock {\em arXiv}, 2021.

\bibitem{Chen20183DCR}
Mingyi Chen, Xuanji He, Jing Yang, and Han Zhang,
\newblock ``3-d convolutional recurrent neural networks with attention model
  for speech emotion recognition,''
\newblock {\em IEEE Signal Processing Letters}, 2018.

\bibitem{Li2019ImprovedES}
Yuanchao Li, Tianyu Zhao, and Tatsuya Kawahara,
\newblock ``Improved end-to-end speech emotion recognition using self attention
  mechanism and multitask learning,''
\newblock in {\em Interspeech}, 2019.

\bibitem{s3prl}
Shu-wen Yang, Po-Han Chi, Yung-Sung Chuang, Cheng-I~Jeff Lai, Kushal Lakhotia,
  Yist~Y Lin, Andy~T Liu, Jiatong Shi, Xuankai Chang, Guan-Ting Lin, et~al.,
\newblock ``Superb: Speech processing universal performance benchmark,''
\newblock {\em arXiv:2105.01051}, 2021.

\bibitem{yoon2018multimodal}
Seunghyun Yoon, Seokhyun Byun, and Kyomin Jung,
\newblock ``Multimodal speech emotion recognition using audio and text,''
\newblock in {\em SLT}, 2018.

\bibitem{wang2020speech}
Jianyou Wang, Michael Xue, Ryan Culhane, Enmao Diao, Jie Ding, and Vahid
  Tarokh,
\newblock ``Speech emotion recognition with dual-sequence lstm architecture,''
\newblock in {\em ICASSP}, 2020.

\bibitem{moma}
Asif Jalal, Rosanna Milner, Thomas Hain, and Roger Moore,
\newblock ``Removing bias with residual mixture of multi-view attention for
  speech emotion recognition,''
\newblock in {\em Interspeech}, 2020.

\bibitem{wise}
Guang Shen, Riwei Lai, Rui Chen, Yu~Zhang, Kejia Zhang, Qilong Han, and Hongtao
  Song,
\newblock ``{WISE: Word-Level Interaction-Based Multimodal Fusion for Speech
  Emotion Recognition},''
\newblock in {\em Interspeech}, 2020.

\bibitem{pepino21_interspeech}
Leonardo Pepino, Pablo Riera, and Luciana Ferrer,
\newblock ``{Emotion Recognition from Speech Using wav2vec 2.0 Embeddings},''
\newblock in {\em Interspeech}, 2021.

\bibitem{actc}
Ziping Zhao, Zhongtian Bao, Zixing Zhang, Nicholas Cummins, Haishuai Wang, and
  Björn~W. Schuller,
\newblock ``{Attention-Enhanced Connectionist Temporal Classification for
  Discrete Speech Emotion Recognition},''
\newblock in {\em Interspeech}, 2019.

\bibitem{li18c_interspeech}
Pengcheng Li, Yan Song, Ian McLoughlin, Wu~Guo, and Lirong Dai,
\newblock ``{An Attention Pooling Based Representation Learning Method for
  Speech Emotion Recognition},''
\newblock in {\em Interspeech}, 2018.

\end{thebibliography}
}

\end{document}